\documentclass[runningheads]{llncs}
\usepackage{graphicx}
\usepackage{amsmath,amssymb} 
\usepackage{tabularx}
\usepackage{color}
\usepackage{url}
\usepackage{algorithm}
\usepackage{hyperref}       
\usepackage{algorithmic}
\usepackage{times}
\usepackage{cite} 
\usepackage{epigraph}
\usepackage[width=122mm,left=12mm,paperwidth=146mm,height=193mm,top=12mm,paperheight=217mm]{geometry}

\renewcommand{\algorithmicrequire}{\textbf{Inputs:}}
\renewcommand{\algorithmicensure}{\textbf{Output:}}

\newcommand{\bs}[1]{\ensuremath{\boldsymbol{#1}}}

\newcommand{\mn}{HTS}
\newcolumntype{Y}{>{\centering\arraybackslash}X}

\begin{document}
\pagestyle{headings}
\mainmatter
\def\ECCV16SubNumber{20}  

\title{Hyperparameter Transfer Learning through Surrogate Alignment for Efficient Deep Neural Network Training}

\titlerunning{Hyperparameter Transfer Learning through Surrogate Alignment}
\authorrunning{Ilija Ilievski, Jiashi Feng}

\author{Ilija Ilievski, Jiashi Feng\\
\texttt{\small ilija.ilievski@u.nus.edu, elefjia@nus.edu.sg}}

\institute{National University of Singapore}

\maketitle

\begin{abstract}
Recently, several optimization methods have been successfully applied to the hyperparameter optimization of deep neural networks (DNNs).
The methods work by modeling the joint distribution of hyperparameter values and corresponding error.
Those methods become less practical when applied to modern DNNs whose training may take a few days and thus one cannot collect sufficient observations to accurately model the distribution.
To address this challenging issue, we propose a method that learns to transfer optimal hyperparameter values for a small source dataset to hyperparameter values with comparable performance on a dataset of interest.
As opposed to existing transfer learning methods, our proposed method does not use hand-designed features.
Instead, it uses surrogates to model the hyperparameter-error distributions of the two datasets and trains a neural network to learn the  transfer function.
Extensive experiments on three CV benchmark datasets clearly demonstrate the efficiency of our method.

\keywords{Hyperparameter Optimization, Transfer Learning, Deep Learning, Image Classification}
\end{abstract}
\section{Introduction}\label{sec:intro}
\vspace{-4mm}
\setlength{\epigraphwidth}{0.65\textwidth}
\epigraph{~Those who cannot learn from history are doomed to repeat it.}{\textit{George Santayana}}
\vspace{-4mm}
Deep neural networks (DNNs) have shown to be very powerful methods and thus have attracted great attention from the computer vision community.
However, their adoption and application are somewhat hampered by their high complexity and in particular the many choices of hyperparameter values (e.g., learning rate, network architectures, activation functions) one must take care of when applying DNN to a new dataset or problem.
Finding appropriate values for the hyperparameters, although essential for achieving good performance, is usually  time consuming and difficult.
Further, complex, state-of-the-art deep learning models (e.g., Residual Networks~\cite{he15}) may take up to several days for training, making the traditional hyperparameter tunning approaches such as grid-search impractical. 
This aroused great interest in developing efficient and systematic hyperparameter optimization approaches~\cite{bergstra2011algorithms, hutter2011sequential, snoek2012practical, swersky2013a, swersky2014freeze, ilievski2016non, bergstra2013making}.
But, even those methods need more than hundred hyperparameter evaluations to find near-optimal hyperparameter values and thus remain impractical for DNN models requiring very long training time. 

A promising approach to address the challenging and highly complex hyperparameter optimization problems is to transfer the knowledge from well-tuned hyperparameters of a DNN evaluated on a source dataset to the hyperparameter optimization for evaluation on a new dataset.
However, the optimal hyperparameter values for different datasets can vary greatly in terms of scale and location.
This makes knowledge transfer for hyperparameter optimization a difficult problem that only few research works have explored. 

Feurer et al.~\cite{feurer2015initializing} address the issue by extracting task and dataset features as metafeatures and use them to initialize the hyperparameter optimization methods.
Bardenet et al.~\cite{bardenet2013collaborative} describe an approach based on surrogate ranking and techniques for collaborative tunning that constructs a common performance surface of the model.
In similar fashion, Yogatama and Mann~\cite{yogatama2014efficient} propose to use the deviations from per-dataset mean as a common representation of the model's performance on two datasets.

Those proposed methods use hand-crafted features for knowledge transfer between hyperparameter optimizations on different datasets. 
However, deep learning models are evidence that hand-designed features are typically inferior to learned features.
With this in mind, we explore a new direction and propose a hyperparameter transfer learning method that has the following two unique features: 
\begin{itemize}
   \item First, it utilizes surrogates to efficiently model the distribution of the validation error given a hyperparameter values and a dataset.
   \item Second, it employs a small neural network to parameterize a knowledge transfer function that maps hyperparameter values from a source dataset to similarly performing hyperparameters values on a target dataset.
\end{itemize}
These unique features enable the proposed method to efficiently optimize the hyperparameters on a dataset of interest while using as much as five times less evaluations, as demonstrated by extensive experiments on three image classification datasets.
We name our method as Hyperparameter Transfer using Surrogates or \mn~for short. 

\section{Related Work}\label{sec:related}

All previous approaches to knowledge transfer for hyperparameter optimization methods depend on hand-crafted features to capture the dataset specific properties. 
In the following, we briefly describe such methods and compare them with our proposed method. 

Yogatama and Mann~\cite{yogatama2014efficient} describe a Bayesian optimization method that maps the validation errors for a set of hyperparameter values on multiple datasets to a common space by scaling the errors with the per-dataset mean and standard deviation.
Their approach is based on the assumption that different datasets produce similar validation errors aside from the location and the scale of the error. 
We argue that the relationship between the validation errors on different datasets is much more complex and cannot be captured solely by the per-dataset deviations.
In contrast, our proposed method provides a transfer function that maps the validation errors on different datasets and thus can be used by any hyperparameter optimization method.  

Feurer et al.~\cite{feurer2015initializing} use dataset metafeatures to compute a dataset distance metric which is then used to transfer high-performing hyperparameter values from ``close'' datasets.
The proposed method is complex and relies on as many as $46$ metafeatures.
Each metafeature carries an assumption about the datasets properties, which may not hold in practice. 
Different from their method, our proposed method learns the dataset properties directly from the performance of the model through training a small neural network.

Both of the above methods employ hand-crafted features to transfer the knowledge gained from hyperparameter optimization on a source dataset to a target dataset.
The \mn~method, learns a function that is able to map hyperparameter values from a source dataset to hyperparameter values offering comparable performance on a target dataset.
To the best of our knowledge, \mn~is the first method to directly learn the parameters of the transfer function that maps hyperparameters across different datasets.

\section{Hyperparameter Transfer using Surrogates}\label{sec:model}
We now describe our  method~\textendash~Hyperparameter Transfer using Surrogates (\mn)~\textendash~for knowledge transfer between hyperparameter optimizations on different datasets. 

\vspace{-1em}
\subsection{Problem Setup}
Consider a deep learning algorithm $\mathcal{A}$ with $k$ configurable hyperparameters collectively denoted as $\bs{x}=\begin{bmatrix}x_1,\ldots,x_k\end{bmatrix}$.
Assume the values of the hyperparameters are constrained to the domain $\mathcal{X}$. 
Hyperparameter optimization aims to find the best hyperparameter configuration $\bs{x}^{*}$ to minimize the validation error of the algorithm $\mathcal{A}$ which is trained with $\bs{x}^{*}$ as its hyperparameter values.
The training and evaluation of $\mathcal{A}$ using the hyperparameter set $\bs{x}$ is considered as evaluation of an expensive function $f:\mathcal{X}\to\mathcal{R}$ that maps a hyperparameter set to a validation error.
We search for $\bs{x}^{*}$ by minimizing the function $f$ with respect to $\bs{x}$.
We focus on learning to transfer the hyperparameter sets that perform well when $\mathcal{A}$ is applied on a source dataset $D_{S}$ to hyperparameter sets that perform well when the same algorithm $\mathcal{A}$ is applied on a target dataset $D_{T}$. 

\vspace{-1em}
\subsection{Method Description}
We define $\Omega_{S}$ as the set of $n$ pairs of hyperparameter configurations and the corresponding validation errors, $\Omega_S \triangleq \{(\bs{x}_i, f_{S}(\bs{x}_i))\}_{i=1}^n$, of the algorithm $\mathcal{A}$ on a source dataset $D_{S}$.
Similarly, we define $\Omega_{T}$ as the set of $(\bs{x},f_{T}(\bs{x}))$ pairs, where $f_{T}(\bs{x})$ is the validation error of the same algorithm $\mathcal{A}$ evaluated on another dataset $D_{T}$ (we call this dataset as the target dataset).

Our goal is to use the best hyperparameter sets from $D_{S}$ to obtain the lowest validation errors on $D_{T}$ as well.
However, the same hyperparameter sets that yielded the lowest validation errors on $D_{S}$ will not necessarily give the lowest validation errors on $D_{T}$ due to the domain shift between the two datasets.
To address this problem, we propose a hyperparameter transfer learning algorithm that automatically adapts the hyperparameter values from one dataset to another.
More concretely, we first define a hyperparameter transfer function $g:\mathcal{X}\to\mathcal{X}$ that maps the hyperparameter configuration $\bs{x}$ to another one $\hat{\bs{x}}$ such that the validation error on the target dataset has a high positive correlation with the validation error on the source dataset.
We parameterize the function $g$ by a small neural network with learnable parameters $\theta$ that can be estimated through minimizing the following  objective function:
\begin{equation}\label{eq:mapf}
   \min_\theta \mathcal{J}_{\theta} \triangleq 1 - {\rm corr}(f_{S}(\bs{x}), f_{T}(\hat{\bs{x}})), \text{ where } \hat{\bs{x}} = g(\bs{x};\theta) 
\end{equation}

A problem with the above (straightforward) approach is the insufficiency of training data for optimizing the neural network $g(\theta)$ since both $\Omega_{S}$ and $\Omega_{T}$ have a small number of elements and populating them with new pairs is very expensive.
We circumvent the problem by substituting $f_{S}$ and $f_{T}$ in Equation~\eqref{eq:mapf} with their surrogate models $S_{S}$ and $S_{T}$ (we use RBF to build the surrogates and  details are deferred to the supplementary material). 
We fit $S_{S}$ using $\Omega_{S}$ and before fitting $S_{T}$ we populate $\Omega_{T}$ by evaluating $m$ Latin hypercube samples (LHS) of $\mathcal{X}$.\footnote{The value of $m$ is a meta-parameter and usually is set between $k$ and $2k$, where $k$ is the number of hyperparameters being optimized.}
Another practical problem is the difficulty of backpropagation through the correlation of the two surrogate functions.
For this, we construct two arrays of $10{,}000$ LHS samples of $\mathcal{X}$ and sort one array according to the value of the sample when evaluated on $S_{S}$ and the other array when evaluated on $S_{T}$.
As a consequence, the $S_{S}$ and $S_{T}$ values of the two sorted arrays has high positive correlation so they are apt substitute for the correlation function. 
Now, the objective function can be simplified to a function that aims to reduce the mean squared distance of hyperparameter values with a same rank in these two arrays.
Accordingly, the neural network training set $\mathcal{M}$ contains LHS samples evaluated on $S_{S}$ as inputs and the LHS samples with same ranked $S_{T}$ values as desired outputs.


\vspace{-1.5em}
\begin{algorithm}
   \caption{Hyperparameter Transfer using Surrogates (\mn)}
   \begin{algorithmic}[1]
      \REQUIRE 
   $\Omega_{S}$ \COMMENT{Set of $n$ pairs of hyperparameters and their evaluation on the source dataset $D_{S}$}\\
      $m$ \COMMENT{Number of Latin hypercube samples (LHS) from $\mathcal{X}$ to initialize $\Omega_{T}$};\\
       $r$ \COMMENT{Transfer step-size};\\
       $t_{\max}$ \COMMENT{Number of evaluations allowed on the target dataset $D_{T}$};\\
      \ENSURE
   $\Omega_{T}$ \COMMENT{Set of $t_{\max}$ pairs of hyperparameters and their evaluations on the target dataset};\\ 
      $g$ \COMMENT{Function that maps $\bs{x}$ s.t.\ $f_{T}(g(\bs{x}))$ has high positive correlation with $f_{S}(\bs{x})$} \\
      \STATE Use $\Omega_{S}$ to approximate the evaluation of a hyperparameter set on the source dataset, $f_{S}$, with a surrogate model $S_{S}$. 
      \STATE Set $\Omega_{T} = \{(\bs{x}_i, f_{T}(\bs{x}_i))\}_{i=1}^m$, where $\bs{x}_{i}$ are $m$ LHS samples from $\mathcal{X}$.
      \STATE Set $t=m$
      \WHILE{$t\leq t_{\max}$}
         \STATE Use $\Omega_{T}$ to approximate the evaluation of a hyperparameter set on the target dataset, $f_{T}$, with a surrogate model $S_{T}$.
         \STATE Set $\mathcal{M}$ as a training set containing LHS samples of $\mathcal{X}$ evaluated on $S_{S}$ as inputs and LHS samples with same ranked $S_{T}$ as desired outputs.
         \STATE Learn a function $g$ by training a neural network on $\mathcal{M}$.
         \STATE After convergence of the neural network, sample $r$ hyperparameter sets from $\Omega_{S}$ with lowest validation errors and set $\Omega_{T}=\Omega_{T} \cup \{(\hat{\bs{x}}_i, f_{T}(\hat{\bs{x}}_i))\}_{i=1}^r$.
         \STATE Set $t=t+r$
      \ENDWHILE
      \STATE Return the set $\Omega_{T}$ and the latest function $g$.
	\end{algorithmic}
	\label{alg:hts}
\end{algorithm}
\vspace{-1.5em}

After convergence of the neural network,\footnote{The convergence time is negligible compared to the duration of one hyperparameter evaluation, the former is in order of seconds while the latter is in order of hours or longer.} we use $g$ to map the $r$ top performing hyperparameter sets from $\Omega_{S}$ and evaluate them using $f_{T}$.\footnote{The value of $r$ is a meta-parameter and can be regarded as the step-size of the transfer learning method. In all our experiments, we set $r$ to $5$.}
We add the $r$ pairs of hyperparameter sets and validation errors $(\hat{\bs{x}}, f_{t}(\hat{\bs{x}}))$ to $\Omega_{T}$ and update the surrogate model $S_{T}$.
We continue by training a new neural network using the updated surrogate model. 
We repeat the cycle until we exhaust the available budget of hyperparameter evaluations on $D_{T}$.
A formal algorithm description is given in Algorithm~\ref{alg:hts}.
We give implementation details, discussions on method variations, and analysis of $m$ and $r$ in the supplementary.

\vspace{-1.3em}
\section{Experiments}\label{sec:experiments} 
\vspace{-2.7em}
\begin{figure}
    \includegraphics[width=1.0\textwidth, height=0.5\textwidth]{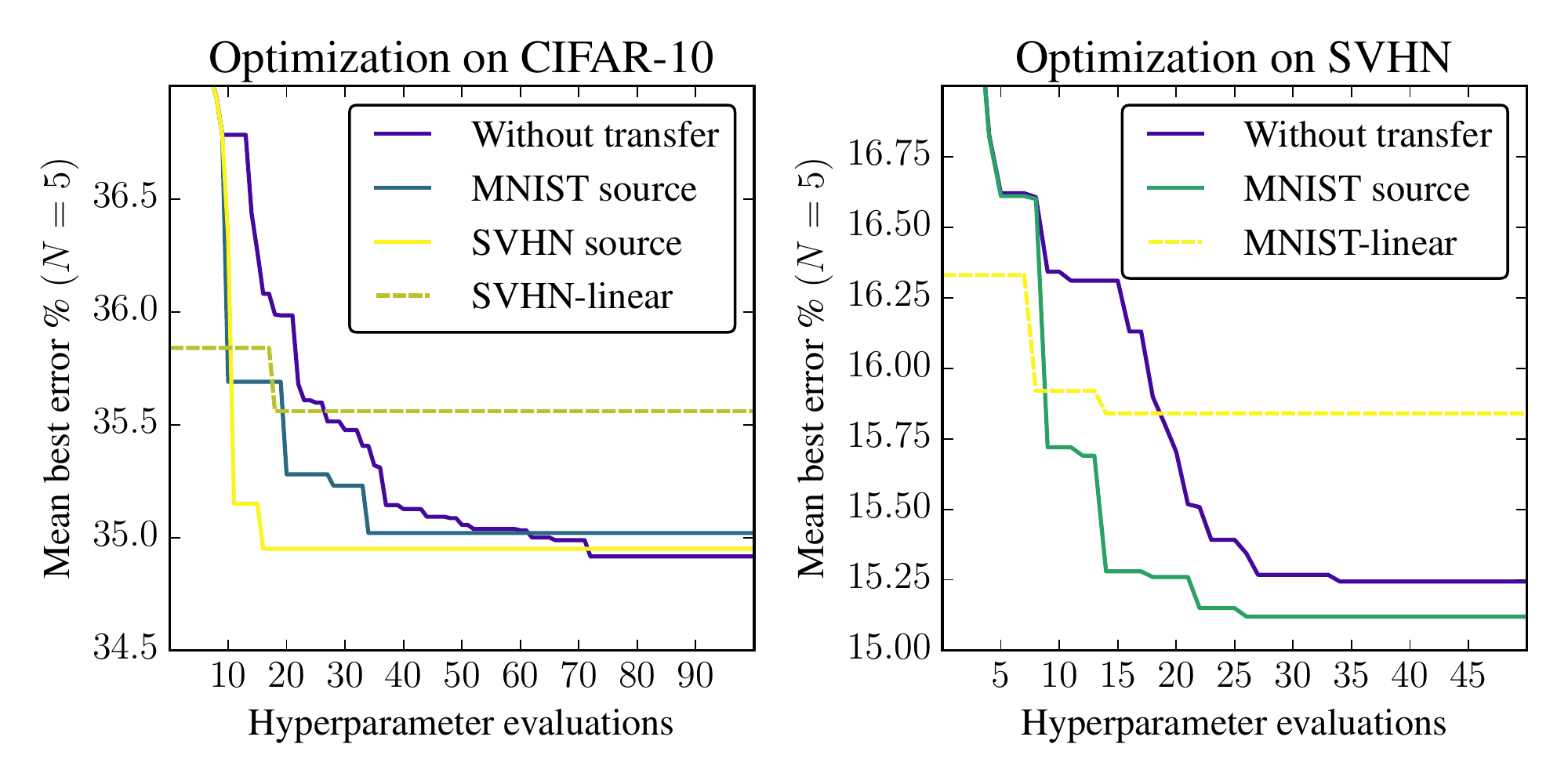}
\vspace{-2.5em}
    \caption{
    Validation error curves for optimizing $8$ hyperparameters of a CNN network on two CIFAR-10 and SVHN.
    Left, we compare the optimization progress on CIFAR-10 when not using a source dataset, when using MNIST as a source dataset, when using SVHN as a source dataset, and finally when using the top-performing SVHN hyperparameter values but without using the mapping function $g$ to adapt them to CIFAR-10, i.e. with linear mapping. 
    Right, we compare the optimization progress on SVHN when not using a source dataset,  when using MNIST as a source dataset, and when mapping the top-performing MNIST hyperparameters without adaptation.
 }
    \label{fig:charts}
\end{figure}
\vspace{-1.3em}

We designed two experiments to demonstrate the efficiency of the \mn~method compared against a baseline method that does not use knowledge transfer.
As a baseline we choose the recently proposed HORD method~\cite{ilievski2016non} since it was shown to obtain state-of-the-art performance on most hyperparameter optimization tasks.

We only compare to this baseline method since we cannot compare to~\cite{feurer2015initializing} as their method is designed to transfer hyperparameters only of a subset of the \textit{same} dataset.
We also cannot compare to~\cite{yogatama2014efficient} as their work deals with transfer between general purpose datasets of small size ($<5{,}000$ samples).
Further, they have not made their method's code public so due to time constraints we leave the reimplementation and subsequent comparison of their method for future work.

In both experiments, we optimize $8$ hyperparameters of a convolutional neural network (CNN).
We use stochastic gradient descent algorithm to train the network, and optimize its learning rate and momentum. 
We also optimize the number of nodes in fully-connected layers and the dropout rate. 
We apply the CNN network on three computer vision benchmark datasets: MNIST, SVHN, and CIFAR-10.
We give details of the CNN, the hyperparameters with their ranges, and datasets details in the supplementary.

In the first experiment, we apply the CNN network on the CIFAR-10 dataset and optimize its hyperparameters.
We compare the optimization progress on CIFAR-10 when not using a source dataset (denoted as ``Without transfer'') against the optimization progress when using MNIST as a source dataset and against when using SVHN as a source dataset. 
Since MNIST contains \textit{grayscale} images of digits, transferring the hyperparameter configurations to the task of classifying \textit{color} images of the everyday objects found in CIFAR-10 requires a complex mapping function. 
Further, simple CNN networks typically achieve a validation error of around $2\%$ on the relatively easy task of classifying MNIST digits, but when applied to the CIFAR-10 dataset they achieve an error of around $37\%$.
Similarly, SVHN contains color images of house numbers, so here simple CNN typically achieves an error of around $16\%$.
This means, the hyperparameter mapping function of \mn~should also learn how to deal with the different scale of the validation error.
Nonetheless, \mn~successfully maps the best found hyperparameters from SVHN to CIFAR-10 and reaches a validation error using five times less evaluations than the baseline method.
The progress is slightly less when using MNIST as source dataset since its task is much easier than SVHN or CIFAR-10. 
Next, we demonstrate the ability of the hyperparameter mapping function to adapt the source hyperparameters to appropriate target hyperparameters. 
For this, we perform an optimization where we sort the hyperparameters according to the validation error on the source dataset and then, in that order, directly evaluate them on the target dataset.
Here we use SVHN as a source dataset and denote the optimization as ``SVHN-linear''.
As expected, this sort of optimization starts with low error, but fails to make progress since its not able to properly adapt the hyperparameters (e.g., number of nodes) to a more complex dataset such as CIFAR-10.

In the second experiment, we apply the same CNN network on the SVHN dataset and optimize the same hyperparameters.
We compare the optimization progress without knowledge transfer against the optimization progress with knowledge transfer from optimization on MNIST.
Similar to experiment~1, the \mn~method achieves low validation errors while using as much as three times less evaluations than the baseline method.
We again demonstrate the effectiveness of the transfer function as it can map hyperparameter values that achieve much lower validation error than the optimization where we directly use the source hyperparameter values.

We illustrate each optimization with Figure~\ref{fig:charts} by reporting the mean best validation error per hyperparameter evaluation over five trials using different random seeds.

\section{Conclusion}\label{sec:conclusion}
We presented \mn, a method for transferring hyperparameter configurations between datasets.
The proposed method efficiently learns to map the top-performing hyperparameter configurations on a source dataset to hyperparameter configurations with comparable relative performance on a target dataset.
The resulting transfer function can be used to transfer the top-performing configurations, but it can also be used to initialize any hyperparameter optimization method and significantly speed up its convergence.
In the future, we plan to evaluate our method on transferring hyperparameter configurations between a broader set of datasets.

\bibliographystyle{splncs}
\bibliography{egbib}
\vspace{9em}
\pagebreak
\setcounter{section}{0}
\renewcommand\thesection{\Alph{section}}
\section{Supplementary}
\subsection{Method Variations and Meta-parameters Settings}

It is important for the initial surrogate $S_{T}$ to be a close approximation of $f_{T}$.
Thus, one might want to use higher number of initial samples $m$, with the expense of spending valuable $f_{T}$ evaluations. 
Further, if the $r$ top performing samples from $\Omega_{S}$ are close to each other in the hyperparameter space, it might be a good idea to also include random samples from $\Omega_{S}$.
The random samples could improve the generalization of the surrogate model and thus produce better training set for subsequent iterations of $g$.

Another possibility is to use \mn~until we have spent only a percentage of the available budget and continue using the surrogate with other hyperparameter optimization algorithm, such as HORD~\cite{ilievski2016non}.
In this case, \mn~can be seen as a hyperparameter optimization initialization method.

\subsection{Experiment Details}
We optimize $8$ hyperparameters of a CNN network, with the following architecture.
The network start with two blocks of: convolutional layer, batch normalization layer, ReLU activation, dropout layer, and a max-pooling layer.
The first block has convolutional layer with $32$ filters and a kernel size of $5\times5$, while the second block a convolutional layer with $64$ filters and a kernel size of $3\time5$.
The max-pooling layers in the first and second block have a kernel size of $3\times3$ and $2\times2$ respectively. 
Finally, the CNN network has two blocks of: fully-connected (FC) layer, ReLU activation, and a dropout layer. 
For training we use the cross-entropy criterion and the stochastic gradient descent (SGD) algorithm.
Due to time constraints and limited computational resources, we train the network only for $10$ epochs.
We tune the learning rate and momentum of the SGD algorithm, the dropout rate of the four dropout layers, and the number of nodes in the two FC layers.
We search for optimal hyperparameter values in the ranges listed in Table~\ref{tbl:hyper}.

\subsection{Implementation Details}

We employ the radial basis function (RBF) with polynomial tail as a generic surrogate model to approximate $f_{S}$ and $f_{T}$.
Specifically, we define a surrogate model of a function with $n$ observations as $S(\bs{x})=\sum_{i=1}^{n} \lambda_{i}\phi(\|\bs{x} - \bs{x_{i}}\|)+p(\bs{x})$. 
Here, $\phi(r)$ is the cubic spline RBF defined as $\phi(r)=r^3$ and $p(\bs{x})$ is the polynomial tail.
The surrogate model parameters are determined by solving the corresponding linear system of equations~\cite{gutmann2001radial}.

Each hyperparameter $x_i$ is constrained to a range $[x_{i,\min}, x_{i,\max}]$.
We normalize each hyperparameter range to a range of $[0, 1]$ before using the hyperparameter sets as neural network training samples. 
Accordingly, we constrain the output of the network to be in the same range $[0, 1]$ and the translate hyperparameter values back to the original range before using as input to $f_{T}$. 
In this way, we avoid the problem of mapping $\bs{x}$ to an undefined or outside range values.

The neural network used to learn the function $g$ consists of one hidden layer with $20k$ neurons and a sigmoid activation function (where $k$ is the number of hyperparameters).
To implement the surrogate model we use the open-source surrogate optimization toolbox pySOT~\cite{pySOT}. 
The neural network for learning $g$ and the CNN for performing the experiments are implemented in the Torch framework~\cite{torch}.

\subsection{Datasets Descriptions}

The MNIST dataset is a popular benchmark dataset for classifying grayscale images of handwritten digits~\cite{lecun1998mnist}.
Following conventional experimental protocol on this dataset, we split the training images into the training set of $50{,}000$ images and the validation set of $10{,}000$ images. 
As a standard preprocessing, we normalized intensity values of all the images by subtracting their mean and dividing by the standard deviation.
Throughout the experiments, we always use the error on the validation set as loss of the objective function we are optimizing.

SVHN is a real-world image dataset obtained from house numbers in Google Street View images~\cite{netzer2011reading}.
It is similat to MNIST (e.g., the images are of small cropped digits), but contains more labeled data ($73{,}257$ train images and $26{,}032$ test images). 
Further, it poses the significantly harder task of recognizing digits and numbers in natural scene images. 
We randomly sample $10{,}000$ training images and use them as a validation set.

The CIFAR-10 dataset consists of $60{,}000$ color images equally divided in $10$ classes: airplane, automobile, bird, cat, deer, dog, frog, horse, ship, and truck~\cite{krizhevsky2009learning}. 
The dataset is split into five training batches and one test batch, each with $10{,}000$ images.
We choose the last training batch as a validation set and use the error on this set to compare the performance of the algorithms.
We also normalized the intensity values of all the images in this dataset by subtracting their mean and dividing by the standard deviation.

\begin{table}[t!]
   \centering
   \caption{Hyperparameters that we optimize with their respective ranges.}
   \setlength{\tabcolsep}{.5em}
   \begin{tabular}{lcc}
      \hline
      Hyperparameter           &   Min.\ value   &   Max.\ value  \\
      \hline
      Learning Rate            &     $0.01$      &    $ 0.40$     \\
      Momentum                 &     $0.60$      &    $ 0.99$     \\
      Dropout Rate $1$         &     $0.00$      &    $ 0.80$     \\
      Dropout Rate $2$         &     $0.00$      &    $ 0.80$     \\
      Dropout Rate $3$         &     $0.00$      &    $ 0.80$     \\
      Dropout Rate $4$         &     $0.00$      &    $ 0.80$     \\
      Nodes in FC Layer $1$    &     $100 $      &    $ 500 $     \\
      Nodes in FC Layer $2$    &     $100 $      &    $ 500 $     \\
   \end{tabular}
   \vspace{-0.5cm}
   \label{tbl:hyper}
\end{table}

\end{document}